\documentclass[10pt,twocolumn,letterpaper]{article}

\usepackage{wacv}              

\usepackage{graphicx}
\usepackage{amsmath}
\usepackage{amssymb}
\usepackage{booktabs}
\usepackage[accsupp]{axessibility}  

\usepackage[pagebackref,breaklinks,colorlinks]{hyperref}

\usepackage[capitalize]{cleveref}
\crefname{section}{Sec.}{Secs.}
\Crefname{section}{Section}{Sections}
\Crefname{table}{Table}{Tables}
\crefname{table}{Tab.}{Tabs.}

\def\wacvPaperID{*****} 
\def\confName{WACV}
\def\confYear{2024}

\begin{document}

\title{Enhancing Surveillance Camera FOV Quality via Semantic Line Detection and Classification with Deep Hough Transform}

\author{Andrew Freeman\\
University of North Carolina at Chapel Hill\\
{\tt\small acfreeman@cs.unc.edu}
\and
Wenjing Shi\\
AWS\\
{\tt\small ntswj@amazon.com}
\and
Bin Hwang\\
AWS\\
{\tt\small hwangbin@amazon.com}
}
\maketitle

\begin{abstract}

The quality of recorded videos and images is significantly influenced by the camera's field of view (FOV). In critical applications like surveillance systems and self-driving cars, an inadequate FOV can give rise to severe safety and security concerns, including car accidents and thefts due to the failure to detect individuals and objects. The conventional methods for establishing the correct FOV heavily rely on human judgment and lack automated mechanisms to assess video and image quality based on FOV. In this paper, we introduce an innovative approach that harnesses semantic line detection and classification alongside deep Hough transform to identify semantic lines, thus ensuring a suitable FOV by understanding 3D view through parallel lines. Our approach yields an effective F1 score of 0.729 on the public EgoCart dataset, coupled with a notably high median score in the line placement metric. We illustrate that our method offers a straightforward means of assessing the quality of the camera's field of view, achieving a classification accuracy of 83.8\%. This metric can serve as a proxy for evaluating the potential performance of video and image quality applications.
\end{abstract}

\section{Introduction}\label{sec:intro}

In today's retail stores, data centers, and self-driving cars, there exists a proliferation of cameras, ranging from just a few to thousands per vehicle or establishment. Prominent examples include AWS data centers, Amazon Go, and self-driving fleets like Waymo and Cruise. The field of view (FOV) of these cameras assumes a pivotal role in enhancing the capabilities of downstream object detection and tracking systems, empowering them to interpret the environment, objects, and human behaviors through computer vision.

In many instances, particularly in surveillance camera deployments, a team of contractors is often tasked with installing a multitude of ceiling-mounted cameras, transmitting continuous video streams to the company's server. Consequently, these cameras can end up misaligned, pointing too high, too low, or even upside-down. Such misconfigurations severely impede the performance of computer vision models, leading to suboptimal results and, at times, outright failures in comprehending the store environment.

In the aforementioned context, one can envision the potential advantages that automated systems would bring to data centers and retail stores, offering the ability to assess whether a camera's field of view meets the quality requirements for accurate downstream applications. This paper addresses this challenge by presenting a method for detecting and classifying prominent straight lines in image collections that exhibit high uniformity, such as store aisles, thereby enabling \textit{field of view classification}.

Our approach builds upon a modification of a previously established model designed for detecting artistic compositional lines in images \cite{deephough}. To gauge the precision of our approach, specifically in retail store scenarios, we conduct evaluations using the publicly available EgoCart dataset \cite{spera2018egocentric}\cite{spera2019egocart}, for which we introduce new labels tailored to our novel line segmentation and classification model, where we use parallel lines of aisles and racks to decide good FOV. Our approach achieves a commendable F1 score of 0.729 for line detection and a great FOV classification accuracy of 83.8\%.

\section{Related Work}

Here, we describe the various methods and goals of existing line detection schemes.

\subsection{Traditional Line Detection}
Early work for detecting lines in images relied on classical computer vision techniques, such as Canny edge detection. Some work using these methods focused on constructing \textit{wireframes}, detecting line segments along planar surfaces, and connecting them to form a schematic view of the 3D space \cite{cannylines}. More recent methods leveraged machine learning to improve the quality of wireframes in challenging scenarios, such as occlusion, curves, and holes \cite{roomwireframe,holes,zhou2019end,elsd}. Asynchronous event cameras have also recently shown promise for further improving wireframe accuracy from high-speed camera motion (e.g., upon a robot) \cite{7605244}.

We distinguish wireframe parsing for scene analysis from line detection aimed at addressing camera shortcomings. We can divide the latter case into the categories of camera calibration (e.g., correcting lens distortion) and camera positioning (e.g., the camera is not aimed at the scene optimally for some applications).


Gong et al. proposed a method for detecting, classifying, and filtering line segments in roadside surveillance cameras \cite{vanishingpointsecurity}. They then extracted the dominant angles and determined the vanishing points of an image. The goal of their work was to automate the detection of poor camera positioning and improve calibration. This work is most similar to the problem we explore in this paper. Gong et al. classified only the directionality of the lines in their images, however, whereas we are interested in the presence and semantic meaning of particular lines, such as the top boundary of a store shelf.


\subsection{Semantic Line Detection}

Lee and Kim et al. introduced the notion of semantic line detection for images \cite{8237612}. Their work focuses on finding photographic \textit{compositional} lines, such as horizon lines and leading lines, by using a CNN-based architecture to predict the four coordinates of these lines in Cartesian space. A primary application the authors explored is composition enhancement, where the location and angle of a detected horizon line may drive the automated correction of an image's rotation and crop region. Unlike the traditional line detection methods detailed above, this approach detects lines that span between two image boundaries.

Jin et al. used a similar CNN approach but proposed an attention-based mechanism to filter the most semantically significant lines from a set of several predictions \cite{jin}. They then refined this approach with a comparison mechanism that seeks to maximize the compositional ``harmony'' of the lines \cite{jin2}. The latter work also offers a great speed improvement over the original.

\subsection{Deep Hough Transform}
The Hough transform is a method for casting image features from Cartesian space into parametric space \cite{Hough}.  Any straight line can be represented in parametric space as the tuple $\langle \theta, r \rangle$. $\theta$ represents the angle of the line in radians, and $r$ represents the distance from the origin to an orthogonal point on the line (where the angle between the line and the origin is 90 degrees). The Hough transform itself works with a voting procedure. If we have an image with reasonably straight lines, we can examine each combination of choices for $\theta, r $, and ``vote" with the intensity values spanned by that line proposal. For standard image processing, this step is best performed after running an edge detector kernel. Then, the parametric representations of the dominant $n$ lines are those $n$ combinations of $\theta, r $, with the highest accumulated response value. Zhao et al. proposed a semantic line detection method that applies a Hough transform to convolutional image features. Their Deep Hough Transform (DHT) model calculates the training loss in parametric space using a $\theta, r $ line representation.


\begin{figure*}[ht]
    \centering
    \includegraphics[width=1.0\linewidth] {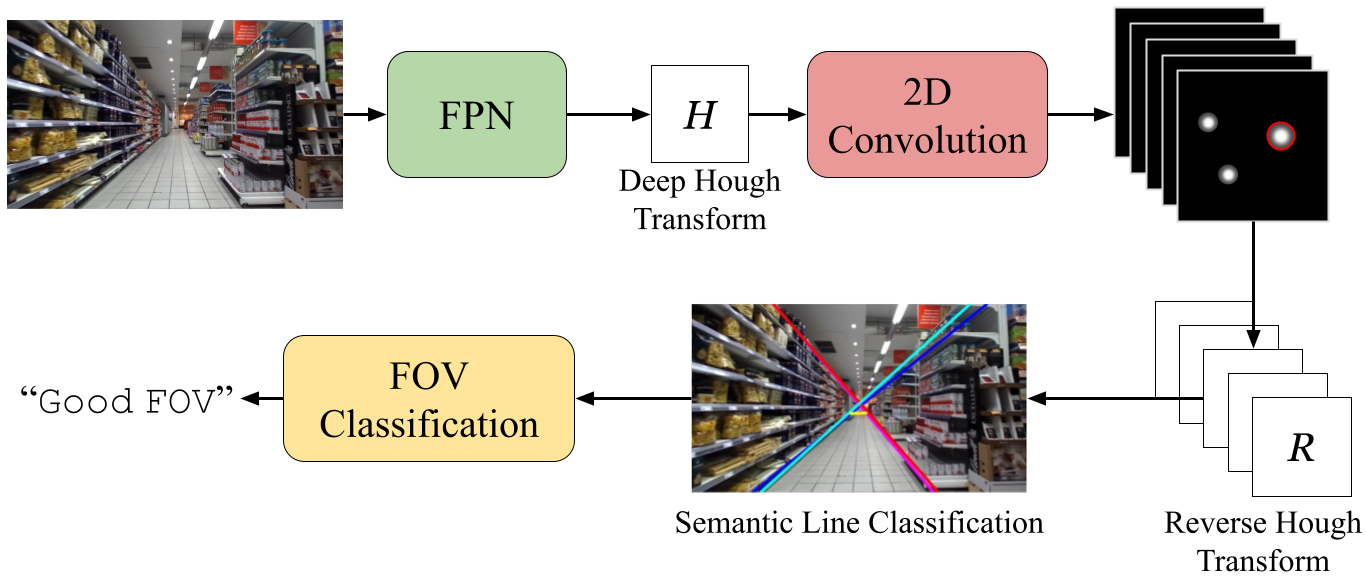}
    \caption{Our modified Deep Hough Transform architecture. We predict lines for multiple semantic classes and take only the single strongest prediction for each class.}
    \label{fig:architecture}
\end{figure*}

\section{Proposed Method}
After a preliminary investigation into the performance and adaptability of the available methods, we chose to modify the original Deep Hough Transform \cite{deephough,eccv2020line} to perform semantic line classification.

\subsection{Standard DHT Architecture}
We here detail our PyTorch \cite{NEURIPS2019_9015} implementation of the Deep Hough Transform architecture introduced by Zhao et al. \cite{deephough,eccv2020line}. The standard DHT model follows an encoder-decoder arrangement. The encoder uses a feature pyramid network (FPN) to extract features from the original image at four resolution levels. At each scale of the FPN output, we run the feature maps through a Hough transform. Corresponding to the four resolution levels, we have four levels of granularity for the number of possible angles ($\#\theta$) and distances ($\#r$) that we examine for these transforms, running a voting procedure for each pixel and $\theta, r$ combination. Batch normalization and rectified linear unit ReLU activations are then applied to the accumulated matrices.

We interpolate these four scales of Hough-space activations to the same resolution and concatenate these matrices together. Finally, we apply one last 2D convolution to get our output. In the standard model, the output is a matrix of dimension $\theta \times r$, and each pixel has a value in the range [0.0, 1.0]. To obtain the $n$ dominant lines, we first isolate the contiguous regions of activated pixels with scikit-image \cite{van2014scikit}. Then, the center coordinate of each region determines its $\theta$ and $r$ values. We simply order these center pixels according to their activation level and take the top $n$ activations that exceed a given threshold (say, 0.1).

\subsection{Model Modifications}

The standard DHT architecture only performs line \textit{detection}, not \textit{classification}. That is, the model may predict any number of lines for a given input image, and there is no in-built scheme for determining each line's semantic meaning. The only method provided for distinguishing a given line from any other is the strength of its activation, which only carries semantic meaning if we limit the training dataset to contain a single type of line (e.g., the horizon line) in each image. For our purposes, we want to predict and classify multiple \textit{specific} lines, such as aisles and the tops of store racks.

To accomplish classification, we modified the last convolutional layer of the DHT architecture to have a variable number of output channels, $N$. Each output channel then represents a single class in our training dataset, and we assume that there can be no more than one instance of each class. Then, instead of taking the $n$ line predictions with the strongest activation as described above, we simply take the strongest single prediction for each class separately (if it exceeds our confidence threshold). Thus, we can perform semantic line classification with few changes to the architecture or inference code. We illustrate our modified architecture in \cref{fig:architecture}.

The original DHT implementation also used a custom CUDA kernel to perform the Hough transform. This kernel was written in C++, requiring prior compilation to be called within the core Python implementation. We reimplemented this kernel in Python with CUDA acceleration through the use of the Numba library, allowing for just-in-time compilation \cite{lam2015numba}. With the low overhead afforded by this library and a more optimized tensor addressing scheme, our implementation decreases the model inference time by up to 12\%.

\begin{figure*}[ht]
    \centering
    \includegraphics[width=1.0\linewidth] {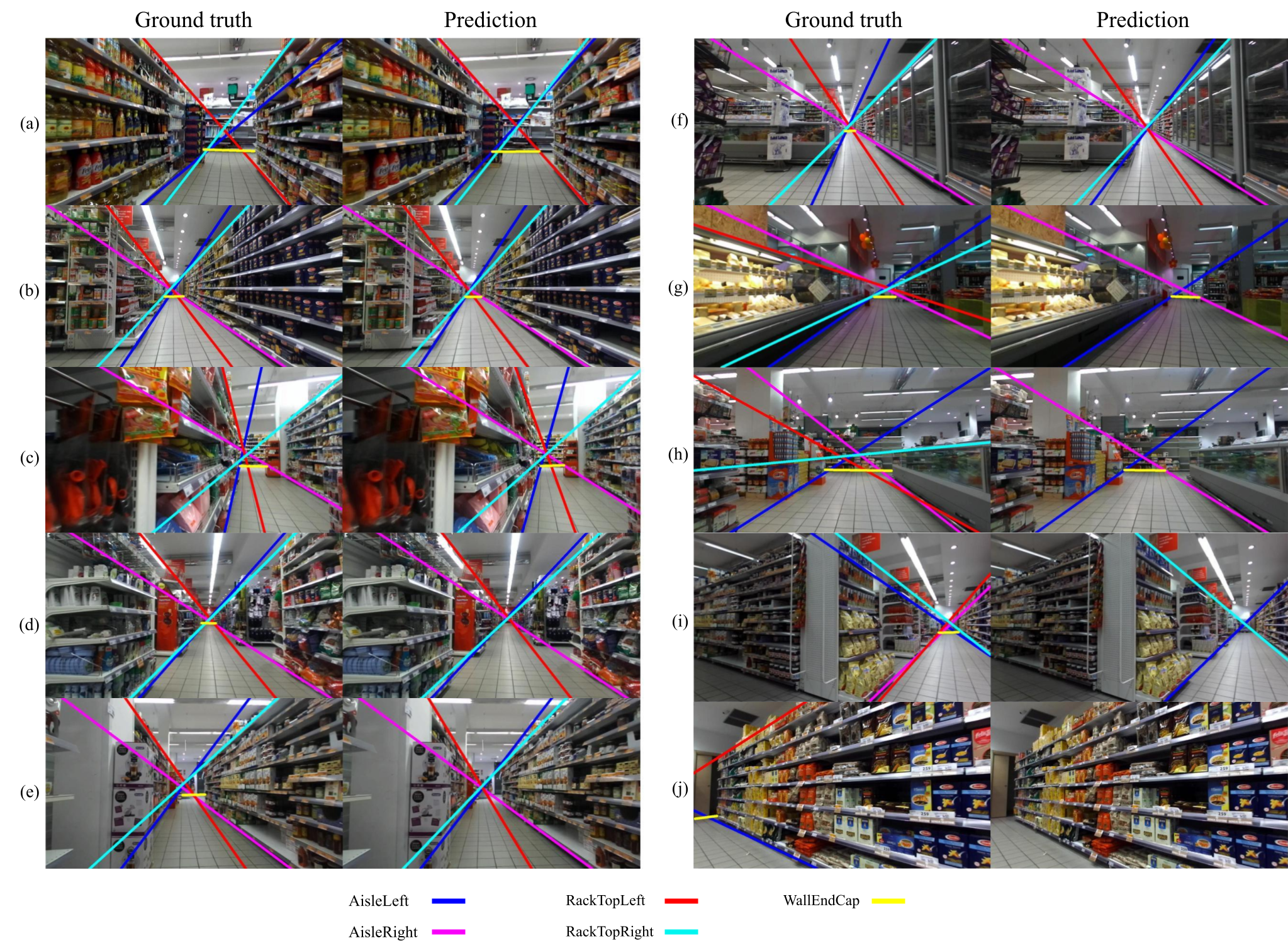}
    \caption{Results on our test dataset. For visual clarity, we illustrate the original coordinates of the \texttt{WallEndCap} class in the ground truth and trim it by its intersection with the \texttt{Aisle} lines in the prediction images. Images (a) through (c) show good predictions with all lines. On images (d) through (f), we fail to predict the small \texttt{WallEndCap} line. Images (g) through (i) represent major failure cases where we fail to predict several lines entirely. Image (j) shows a heavily skewed camera angle, with no classes predicted.}
    \label{fig:results_matrix}
\end{figure*}

\section{Model Evaluation}

\subsection{Dataset}
As there was no existing dataset for semantic line classification, we created our own labels for the EgoCart\cite{spera2018egocentric}\cite{spera2019egocart} image dataset. The dataset contains 19,531 RGB images captured in a real-world grocery store, including varying customer viewing angles from 9 videos. We selected images by uniformly sampling every 20 frames, resulting in a total of 977 images. Using these sampled frames, we annotated the following classes: 
\begin{itemize}
    \item \texttt{AisleLeft}: the intersection of the left store shelves with the floor
    \item \texttt{AisleRight}: the intersection of the right store shelves with the floor
    \item \texttt{RackTopLeft}: the upper boundary of the left store shelves
    \item \texttt{RackTopRight}: the upper boundary of the left store shelves
    \item \texttt{WallEndCap}: the intersection of the far wall with the floor
\end{itemize}

We show examples of these ground truth labels in \cref{fig:results_matrix}.

\subsection{Backbone and Pre-Training}
We used ResNet50 \cite{7780459} as our encoder backbone, with the default pre-trained weights from PyTorch on AWS SageMaker. We did not employ transfer learning from the DHT model weights, as it increased the time to model convergence and offered no benefit to our model's performance. 

\subsection{Evaluation Metric}
The DHT paper introduces the robust Euclidean and Angular (EA) score to quantify the similarity of a predicted line, $l_p$, and a ground truth line, $l_g$. The EA score formula given is

\begin{equation}
    S = \biggl(\bigl(1 - \frac{\theta(l_p, l_g)}{\pi / 2}\bigr)\cdot\bigl(1-D(l_p,l_g)\bigr)\biggr)^2,
\end{equation}

where $\theta(l_p, l_g)$ is the angle between the two lines (in radians) and $D(l_p,l_g)$ is the Euclidean distance between the midpoints of the lines \cite{deephough}. The authors use a bipartite graph to match the candidate line predictions to the corresponding ground-truth lines.

Since we perform line \textit{classification} and since our ground truth contains no more than one line per class, we obviate the need for line matching. Rather, if a class has any activation exceeding a given threshold, the strongest activation for that class constitutes our prediction. We clarify the following special cases to our calculation of the EA score:

\begin{equation}
    S =\begin{cases} 
      0 &  \exists l_p, \exists! l_g \\
      0 &  \exists! l_p, \exists l_g \\
      1 &  \exists! l_p, \exists! l_g \\
   \end{cases}
   \label{eqn:2}
\end{equation}

By this, we heavily penalize false positives and false negatives in the calculation of our loss function, so that we can more accurately determine the binary presence of a line for a given class.

\subsection{Training}

We randomly sampled 668 images (68\%) from our dataset. Of these images, we used 80\% for training and 20\% for validation. The remaining 309 images constituted our test dataset. Images were rescaled to a $1200\times 1200$ resolution and input in batches of 12. We used the Adam optimizer \cite{KingBa15} with a learning rate of 0.0002, momentum of 0.9, and gamma of 0.1. We stopped training after 100 epochs.

For our modified Deep Hough model, we set $\#r = 150$ and $\#\theta = 150$. We set our classification threshold at 0.01, meaning that some pixel in a given class' prediction matrix must have an activation greater than 0.01 for that class to be deemed present in the image.

\subsection{Results}\label{sec:results}
On our validation dataset during training, we achieved 0.729 on the F1 metric and 0.287 on the F1 metric with $95\%$ confidence. For comparison, the original DHT model achieved F1 scores of 0.786 and 0.719 on the SEL \cite{8237612} and NKL \cite{deephough} datasets, respectively, which only involve line detection (not classification).

\begin{table}
    \centering
    \resizebox{\columnwidth}{!}{%
    \begin{tabular}{|l|cccc|}
                             & Precision & Recall & Accuracy & F1 \\
                             \hline
        \texttt{AisleLeft}   & 0.96 & 0.80 & 0.84 & 0.88 \\
        \texttt{AisleRight}  & 0.93 & 0.80 & 0.83 & 0.86 \\
        \texttt{RackTopLeft} & 0.92 & 0.56 & 0.65 & 0.69 \\
        \texttt{RackTopRight}& 0.96 & 0.63 & 0.72 & 0.76 \\
        \texttt{WallEndCap}  & 0.94 & 0.54 & 0.62 & 0.69 \\
    \end{tabular}
    }
    \caption{Test results for each class in our dataset. We achieve high precision for all classes but low recall for \texttt{RackTopLeft}, \texttt{RackTopRight}, and \texttt{WallEndCap}. }
    \label{tab:results_table}
\end{table}

We break down the per-class results of our evaluation on the test dataset in \cref{tab:results_table}, showing high precision in all classes but low recall in the \texttt{RackTop} and \texttt{WallEndCap} classes. Although some classes may not always have a line prediction reaching our confidence threshold, these results show that the \textit{presence} of a line prediction means that we can be highly confident about the spatial \textit{placement} of that line (based on the precision results). As further evidence for this claim, we measured a mean EA score of 0.650 and a median EA score of 0.910 across all line predictions on our test dataset. Our mean EA score is much lower due to the presence of incorrect line detections (see \cref{eqn:2}).

\begin{figure*}[ht]
    \centering
    \includegraphics[width=1.0\linewidth] {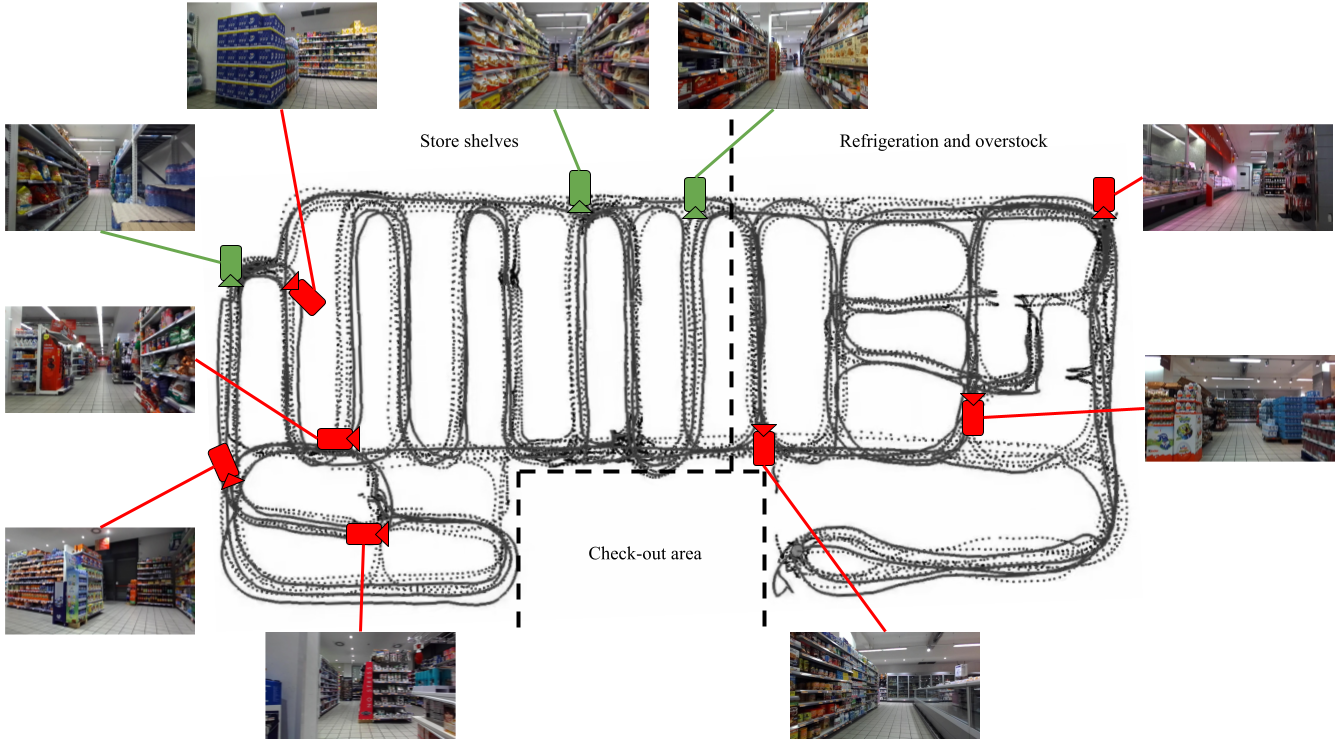}
    \caption{Example camera positions illustrated on the store layout from \cite{spera2019egocart}. Red camera icons denote a \texttt{bad} FOV ground truth classification according to the criteria in \cref{sec:fov_eval}, while green icons denote a \texttt{good} FOV.}
    \label{fig:store_layout}
\end{figure*}

We offer examples from our test dataset in \cref{fig:results_matrix}. We achieve strong results on images with two clearly-defined sets of store shelves, where the camera has a clear view of the end of the aisle. For example, \cref{fig:results_matrix} (a) through (c) show such images where all five lines are predicted and placed with high precision. However, we often do not predict the \texttt{ WallEndCap} class, as in \cref{fig:results_matrix} (d) through (f). This is not surprising, since the length of that line tends to be very short compared to the other classes, so its features are less prominent. More ambiguous images, as in \cref{fig:results_matrix} (g) through (i), lead to multiple classes missing predictions altogether. These failures tend to occur when the primary object on one or more sides of the image is not a standard shelving unit. For example, refrigeration units are common in the cross sections between aisles, as in \cref{fig:results_matrix} (g) and (h), and these lines are often not predicted. Missing line detections can also occur when the camera is at a heavily skewed angle and not pointing directly down an aisle, as in \cref{fig:results_matrix} (i) and (j).

\section{FOV classification}
We do not view these line detection failures as an outright negative result, however. As noted above, such failures predominately occur when the camera is positioned such that its view is atypical. Our work chiefly aims to detect cameras with a poor field of view (FOV) for surveillance applications, to optimize the positioning of many cameras, and to ensure the availability of high-quality quality images for downstream vision tasks. Therefore, if line detection \textit{failure} carries a semantic meaning in our model, we can make informed decisions about the quality of a camera's FOV. Here, we describe an experiment for FOV classification and discuss the implications.

\subsection{Evaluation}\label{sec:fov_eval}
We devise a scheme for a binary classification of store camera FOV. An image can either have a \texttt{good} or \texttt{bad} FOV. An image is said to have a \texttt{good} FOV when the following criteria are met:
\begin{itemize}
    \item There is one clear aisle visible (the camera is not pointed between two aisles)
    \item There are full-height shelving racks on both the left and right sides
    \item The tops of both sets of shelving racks are visible
    \item The length of the aisle is roughly centered in line with the camera (the right shelves proceed about as long as the left shelves)
\end{itemize}
While the cameras in the EgoCart dataset are positioned at a human eye level rather than ceiling-mounted, the definitions of \texttt{good} and \texttt{bad} FOV are transferable to the surveillance-style camera setups. For our ground truth, we hand-labeled the test dataset with these criteria. $62.5\%$ of the ground truth test images were considered \texttt{ good} FOV, while $37.5\%$ were \texttt{bad}. We provide examples of these FOV classes in \cref{fig:store_layout}.

After understanding the typical failure cases of our line classification model (\cref{sec:results}), we devised a remarkably simple criterion to predict FOV: If our modified DHT model detects two \texttt{Aisle} lines and two \texttt{RackTop} lines, regardless of their exact placement, we classify the image as having \texttt{good} FOV.

\subsection{Results}
With the above criteria, we classified the FOV of images in our test dataset with \textbf{83.8\% accuracy}. Since our semantic line classification model fails predictably in certain cases, we see that such failures carry \textit{semantic meaning}. That is, if \texttt{Aisle} lines are not predicted, the camera angle is likely to be skewed; if \texttt{RackTop} lines are not predicted, the camera may be too close to the end of an aisle or not be positioned between two sets of shelves.

Our method reveals a direct metric for camera pose \textit{quality}. We do not require multi-view analysis or depth data to make this assessment, as we do not directly calculate the camera pose. Rather, we use the quality of our semantic line classification prediction as a proxy for the performance of higher-level applications. In the scenario we proposed in \cref{sec:intro}, for example, one can see that a shelf inventory tracking model will perform better if we can ensure its input is high quality and consistent between various cameras.

\section{Future Work}
To continue this work, we will evaluate the efficacy of our model and classification scheme on ceiling-mounted surveillance camera data. While such a dataset of grocery store images would be the most direct extension, we expect the line classification model to perform well in any consistent interior setting, such as data centers, warehouses, and libraries.

Additionally, we will investigate the impact of FOV classification on higher-level vision applications, such as object tracking and semantic segmentation. To simulate technicians correcting \texttt{bad} FOV camera poses, we will require a dataset with images from multiple camera views at the same physical locations. Then, we may fully evaluate the effect of FOV correction on the performance of these applications.

\section{Conclusion}
In summary, this paper introduces novel enhancements to the Deep Hough Transform model, enabling it to not only detect but also classify semantic lines. Through rigorous testing on real-world images sourced from the EgoCart dataset, we have achieved great accuracy in semantic line classification with an F1 score of 0.729. Furthermore, our study illustrates the practical application of semantic line classifications in the context of the camera Field of View (FOV) classification, attaining an impressive 83.8\% accuracy rate. This novel approach enables the estimation of image quality in relation to high-level vision applications, offering an automated means to identify cameras in need of manual pose adjustment or recalibration. In doing so, we extend the utility of semantic line detection models beyond their traditional role as aids for photographic composition, showcasing their relevance and effectiveness in broader domains.

{\small
\bibliographystyle{ieee_fullname}
\bibliography{egbib}
}

\end{document}